\documentclass{article}

\usepackage[final]{neurips_2021}

\usepackage[utf8]{inputenc} 
\usepackage[T1]{fontenc}    
\usepackage{hyperref}       
\usepackage{url}            
\usepackage{booktabs}       
\usepackage{amsfonts}       
\usepackage{nicefrac}       
\usepackage{microtype}      
\usepackage{xcolor}         
\usepackage{graphicx}
\usepackage{comment}

\title{Scalable Geometric Deep Learning \\ on Molecular Graphs}

\author{%
  Nathan C. Frey\\
  MIT\\
  \texttt{ncfrey@mit.edu} \\
  \And
    Siddharth Samsi\\
  MIT\\
  \texttt{ssamsi@mit.edu} \\
    \And
    Joseph McDonald\\
  MIT\\
  \texttt{jpmcd@mit.edu} \\
  \And
   Lin Li\\
  MIT\\
  \texttt{lin.li@ll.mit.edu} \\
  \And
    Connor W. Coley\\
  MIT\\
  \texttt{ccoley@mit.edu} \\
    \And
    Vijay Gadepally
    \thanks{DISTRIBUTION STATEMENT A. Approved for public release. Distribution is unlimited. This material is based upon work supported by the Under Secretary of Defense for Research and Engineering under Air Force Contract No. FA8702-15-D-0001. Any opinions, findings, conclusions or recommendations expressed in this material are those of the author(s) and do not necessarily reflect the views of the Under Secretary of Defense for Research and Engineering. © 2021 Massachusetts Institute of Technology. Delivered to the U.S. Government with Unlimited Rights, as defined in DFARS Part 252.227-7013 or 7014 (Feb 2014). Notwithstanding any copyright notice, U.S. Government rights in this work are defined by DFARS 252.227-7013 or DFARS 252.227-7014 as detailed above. Use of this work other than as specifically authorized by the U.S. Government may violate any copyrights that exist in this work.} \\
  MIT\\
  \texttt{vijayg@mit.edu} \\

}

\begin{document}

\maketitle

\begin{abstract}
  Deep learning in molecular and materials sciences is limited by the lack of integration between applied science, artificial intelligence, and high-performance computing. Bottlenecks with respect to the amount of training data, the size and complexity of model architectures, and the scale of the compute infrastructure are all key factors limiting the scaling of deep learning for molecules and materials. Here, we present \textit{LitMatter}, a lightweight framework for scaling molecular deep learning methods. We train four graph neural network architectures on over 400 GPUs and investigate the scaling behavior of these methods. Depending on the model architecture, training time speedups up to $60\times$ are seen. Empirical neural scaling relations quantify the model-dependent scaling and enable optimal compute resource allocation and the identification of scalable molecular geometric deep learning model implementations. 
\end{abstract}

\section{Introduction}

Many blockbuster results in deep learning are enabled by immense scale. Successes in natural language processing, computer vision, and applied scientific domains like protein folding rely on models that effectively scale to billions of parameters and tens or hundreds of GPUs. A distinguishing feature of these deep-learning-driven results compared to "vanilla" machine learning methods is their "unreasonable effectiveness" and ability to leverage increasing resources (data, model complexity, compute infrastructure) without quickly hitting asymptotic limits on performance [\citenum{Sejnowski30033}]. Furthermore, large models hold the promise of emergence [\citenum{bommasani2021opportunities}], wherein behaviors arise that cannot be anticipated by studying smaller models. To achieve this level of impact in scientific domains such as chemistry and materials science, there is a need to identify and investigate scalable model architectures that operate on molecular graphs.

\begin{figure}
  \centering
  \includegraphics[width=\columnwidth]{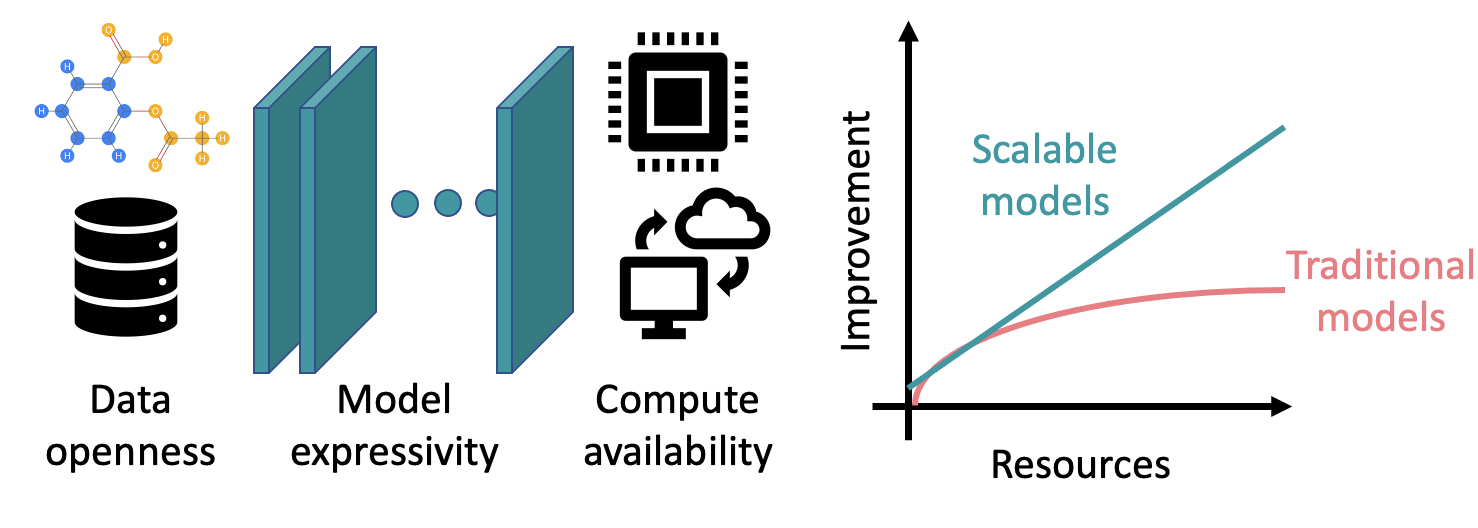}
  \label{bottlenecks}
  \caption{Bottlenecks to scaling geometric deep learning on molecular graphs.}
\end{figure}

In the field of molecular machine learning (ML), there is a staggering number of model architectures and methods to choose from [\citenum{Butler2018}], but most are never tested beyond the low-resource limit [\citenum{pappu2020making}] (small benchmark datasets, small models, and single-GPU training). This makes it difficult to identify which methods are robust enough to be production ready for use cases beyond standard benchmarks [\citenum{dwivedi2020benchmarking}].  For most physical scientists, it is difficult to justify the engineering effort and additional experimentation required to apply an existing method to their problem of interest. Moreover, in low-resource settings, deep learning methods struggle to outperform baselines of simple models (i.e., random forests, logistic regression) and human-crafted representations (i.e., chemical fingerprint vectors) on molecular ML tasks [\citenum{pappu2020making, C8SC00148K}]. 
To scale molecular ML beyond the low-resource limit, Figure~\ref{bottlenecks} summarizes some of the key driving factors: training data availability, expressivity of models for learning molecular representations, and the availability of compute resources needed for hyperparameter optimization, architecture search, and model training. The availability of large-scale molecular and materials databases provides an unprecedented wealth of heterogeneous datasets and learning tasks for molecular modeling. Due to the natural graph structure in condensed matter systems such as molecules, polymers, and crystalline materials [\citenum{Kearnes_2016}], geometric deep learning models invoke the necessary geometric priors for representing molecular data manifolds [\citenum{7974879, bronstein2021geometric}], in the same way that transformers and convolutional layers are natural network topologies for language and images, respectively. However, there still remains a need to investigate the scaling behavior of these models with respect to the available compute infrastructure. 

Here, we investigate the scaling behavior of molecular graph neural networks via distributed training on many GPUs. We present the \textit{LitMatter} template, a lightweight framework for rapid prototyping and scaling deep learning on molecular graphs. We train four state-of-the-art GNN models using up to 416 GPUs. We find that different standard model implementations achieve training time speedups as dramatic as $60\times$ and as low as $3.7\times$, when scaling the number of GPUs from 2 to 416, depending on the model's ability to leverage increasing compute resources. Neural scaling relations [\citenum{hestness2017deep, bahri2021explaining, henighan2020scaling}] are proposed to quantify the model-dependent scaling. These empirical scaling relations can be used to identify models that benefit from increasing compute and to optimally allocate compute resources for efficient training of large models. We envision that the results presented here will help to "bridge the gaps" between high-performance computing, AI, and molecular science. 

\section{LitMatter}

\begin{figure}
  \includegraphics[width=8cm]{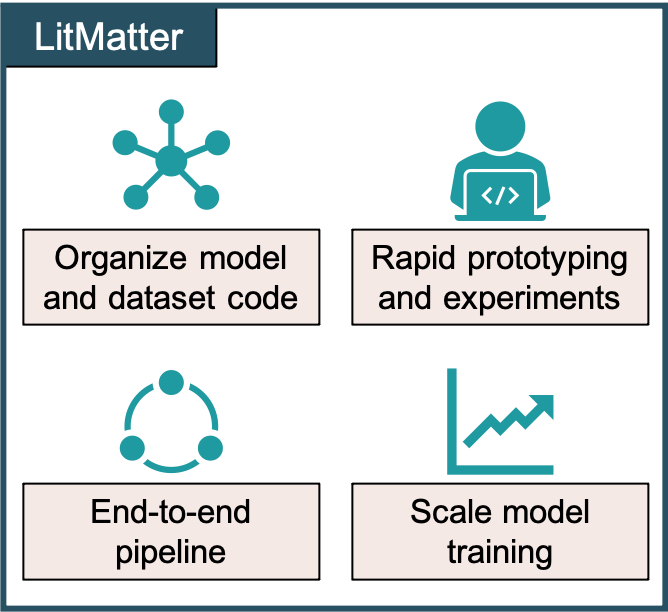}
  \centering
  \label{litmatter}
  \caption{LitMatter: a template for rapid experimentation and scaling deep learning models on molecular graphs.}
\end{figure}

To facilitate easy distributed training of GNNs and other model architectures relevant for molecular and materials science, we present \textit{LitMatter}\footnote{https://github.com/ncfrey/litmatter}, a lightweight template for rapid experimentation and scaling. The main features of \textit{LitMatter} are summarized in the schematic in Figure~ \ref{litmatter}. \textit{LitMatter} uses \textit{PyTorch Lightning} [\citenum{falcon2019pytorch}] to organize code so that engineering best practices are enabled by default and to reduce the overhead required to enable distributed training. \textit{LitMatter} uses the the Distributed Data Parallel [\citenum{li2020pytorch}] accelerator for multi-GPU training. The template is designed to span an entire deep learning workflow, from rapid prototyping to end-to-end training/inference pipeline on hundreds of GPUs. Arbitrary PyTorch models and datasets/dataloaders can be "injected" into the template; and training can be launched interactively in a Jupyter notebook, through the command line, or scheduled through a SLURM job \emph{without modifying or adding any code}. Using the \textit{LitMatter} template allows any researcher to run the experiments presented here, or investigate the scaling behavior of other models of interest.

\section{Scaling Experiments}

We are interested in identifying GNN model architectures that leverage increasing amounts of compute. For reproducibility and simplicity, we choose four state-of-the-art GNN architectures implemented in \textit{PyTorch Geometric} [\citenum{Fey/Lenssen/2019}] - DimeNet [\citenum{klicpera2020directional}], NNConv [\citenum{Gilmer2017}], SchNet [\citenum{schutt2017schnet}], and Principal Neighborhood Aggregation (PNA) [\citenum{Corso2020}]. DimeNet, NNConv, and SchNet were trained to predict quantum chemical properties of small molecules from the QM9 [\citenum{ramakrishnan2014quantum}] dataset, while PNA was trained on regression tasks for 250K compounds from the ZINC15 [\citenum{Sterling2015}] database. All model implementations and datasets were taken directly from \textit{PyTorch Geometric}. All hyperparameters were set to the values provided in the \textit{PyTorch Geometric} examples. 

Multi-node, multi-GPU training was achieved using the \textit{LitMatter} framework. Each model was trained for a fixed number of epochs (200 for DimeNet and 1000 for all other models) and with a fixed batch size of 128 on a variable number of GPUs. The total training time was logged for each model on 2, 4, 8, 16, 32, 64, 128, 256, 364, and 416 GPUs (with the exception of PNA on 416 GPUs). Then the training time per epoch was calculated for each model for each training run. Models were trained on NVIDIA Volta V100 Tensor Core GPUs.

\begin{figure}
  \includegraphics[width=0.6\columnwidth]{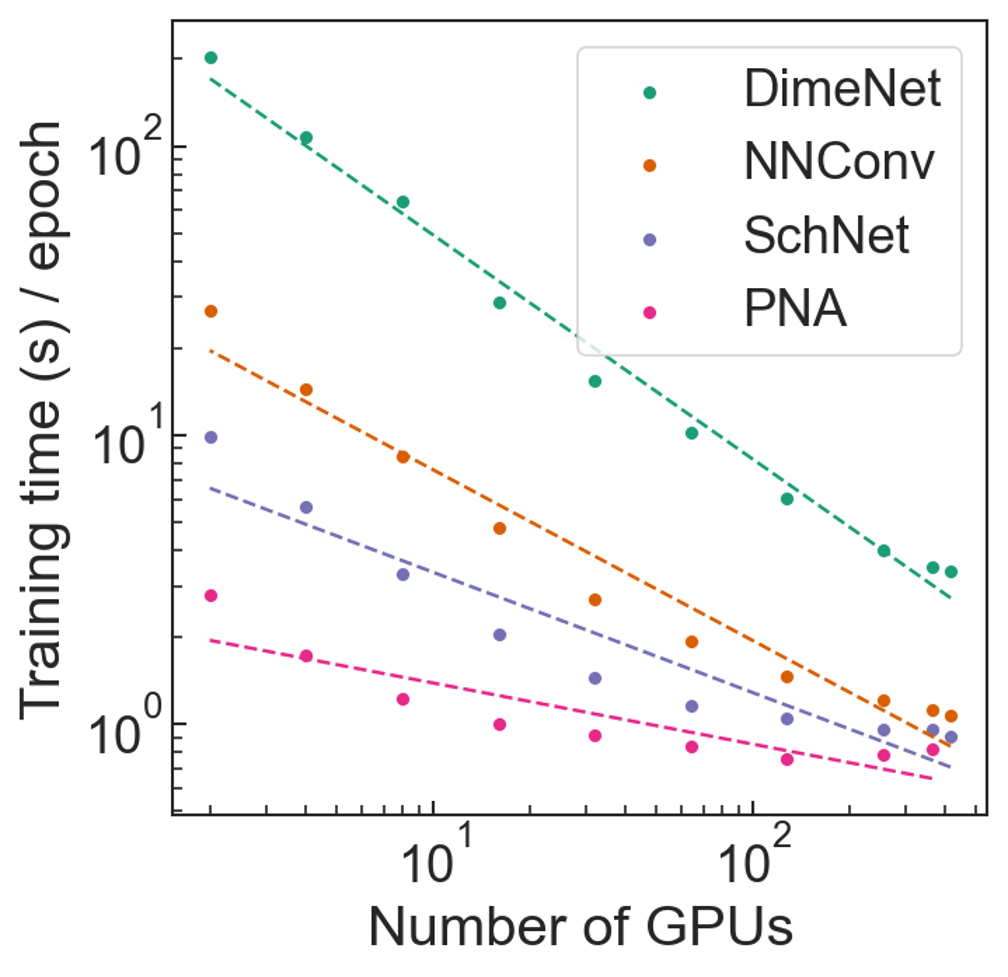}
  \centering
  \label{scaling}
  \caption{Training time per epoch versus number of GPUs for various GNN models.}
\end{figure}

The results of the scaling experiments are shown in Figure~ \ref{scaling}. The most extreme speedup is a 50x performance improvement for DimeNet, which has a training time of over 200 s / epoch on two GPUs that decreases to less than 4 s / epoch on 416 GPUs. NNConv training time decreases from 27 s / epoch (2 GPUs) to 1.08 s / epoch (416 GPUs) for a 25x speedup. SchNet's training time improves from 9.9 s / epoch (2 GPUs) to 0.91 s / epoch (416 GPUs) for an 11x speedup. PNA shows the least dramatic performance improvement, with a training time of 2.8 s / epoch on 2 GPUs and a minimum training time of 0.76 s / epoch on 128 GPUs, corresponding to a 3.7x speedup. PNA is the only model that does not exhibit a monotonic decreases in training time per epoch with increasing number of GPUs, due to inter-node communication bottlenecks at higher counts of compute nodes. For all four models, we observe that there is a clear inflection point at 32 GPUs where allocating additional compute yields diminishing returns. This could reflect a limitation of the dataset sizes (\char`\~100K samples for QM9 and ~250K for ZINC15), bottlenecks in the inter-node communications, and/or constraints on model complexity.

\section{Power Law Scaling}
To further quantify the scaling behavior of the GNN architectures, we fit the scaling data to power laws of the form
\begin{equation}
    t = \alpha n^{-\beta}
\end{equation}
where $t$ is the training time per epoch and $n$ is the compute (number of GPUs). $\alpha$ is a constant and $\beta$ is the scaling exponent that determines the slope of the power law scaling. 

The DimeNet data exhibits almost a perfect power law fit ($R^2=0.99$) with $\beta=0.78$. The magnitude of the scaling exponent reflects the speedup achievable for DimeNet. Table \ref{table1} summarizes the model size, scaling exponent, and goodness of fit for the power law fits for all GNN architectures. 

The decreasing magnitude of $\beta$ going from DimeNet to NNConv, SchNet, and PNA indicates the relative speedup each model achieves through multi-GPU training. All power law fits have $R^2>=0.9$ except PNA, which shows a deviation from power law scaling above 128 GPUs due to the previously mentioned inter-node communication bottleneck.

\begin{table}
  \caption{Power law exponents}
  \label{table1}
  \centering
  \begin{tabular}{llll}
    \toprule
    Model     & Number of Parameters
    & $\beta$ & $R^2$ \\
    \midrule
    DimeNet & $2.1*10^6$  & 0.78 & 0.99     \\
    NNConv & $6.2*10^5$ & 0.59 & 0.96 \\
    SchNet & $4.6*10^5$ & 0.42 & 0.90 \\
    PNA & $6.8*10^5$ & 0.21 & 0.79 \\
    \bottomrule
  \end{tabular}
\end{table}

\section{Discussion}
In this paper we presented the \textit{LitMatter} framework to overcome difficulties in scaling molecular deep learning, enabling researchers to rapidly experiment with and scale deep learning methods on molecular graphs. We showed the scaling behavior of molecular graph neural networks with respect to compute. DimeNet, a GNN with continuous equivariance to rotations (the SO(3) group), exhibits the most dramatic reduction in training time with increased compute. Because equivariant networks like DimeNet use spherical Bessel functions and spherical harmonics to construct representations, they have significant memory requirements and can be difficult to train. SOTA benchmark results [\citenum{klicpera2020directional}] and the attractive properties of equivariance suggest that equivariant GNNs may be the "ideal" network topology for molecular ML. The scaling behavior of GNNs was quantified by fitting the data to power laws and extracting the scaling exponents. The empirical scaling relations succinctly encapsulate the results of our experiments and can be used in a practical setting to optimally allocate compute resources during training.  Integrating our scaling results with previous work on GNN pre-training strategies [\citenum{hu2020strategies}] and demonstrations of the sample-efficiency and ability of large models to avoid overfitting through early stopping [\citenum{bommasani2021opportunities, henighan2020scaling}] suggests a recipe for large GNNs that are trained at scale and fine-tuned to downstream molecular ML tasks. Next steps for this work include investigating molecular graph neural scaling behavior with respect to other limiting factors (dataset and model size) and identifying a candidate model architecture with optimal scaling behavior that can serve as a molecular graph "foundation model." 

\begin{ack}
The authors acknowledge the MIT SuperCloud [\citenum{reuther2018interactive}] and Lincoln Laboratory Supercomputing Center for providing HPC and consultation resources that have contributed to the research results reported within this paper/report. The authors acknowledge the MIT SuperCloud team: William Arcand, David Bestor, William Bergeron, Chansup Byun, Matthew Hubbell, Michael Houle, Mike Jones, Jeremy Kepner, Anna Klein, Peter Michaleas, Lauren Milechin, Julie Mullen, Andrew Prout, Albert Reuther, Antonio Rosa, and Charles Yee. The authors also wish to acknowledge the following individuals for their contributions and support: Bob Bond, Allan Vanterpool, Tucker Hamilton, Jeff Gottschalk, Tim Kraska, Mike Kanaan, Charles Leiserson, Dave Martinez, John Radovan, Steve Rejto, Daniela Rus, Marc Zissman.

\end{ack}


\setcitestyle{numbers}
\bibliographystyle{unsrtnat} 
\bibliography{bib}

\end{document}